\documentclass[default,iicol]{sn-jnl}

\usepackage[ruled,linesnumbered]{}

\usepackage{amsmath,amsfonts,bm}

\def\eqref#1{equation~\ref{#1}}

\def\1{\bm{1}}

\DeclareMathAlphabet{\mathsfit}{\encodingdefault}{\sfdefault}{m}{sl}
\SetMathAlphabet{\mathsfit}{bold}{\encodingdefault}{\sfdefault}{bx}{n}

\usepackage{booktabs}
\usepackage{subfigure}
\usepackage{hyperref}
\usepackage{color,soul}
\usepackage{algorithm} 
\usepackage{soul}
\usepackage{comment}

\jyear{2022}%

\theoremstyle{thmstyleone}%
\theoremstyle{thmstyletwo}%

\theoremstyle{thmstylethree}%

\begin{document}

\title[Article Title]{\textbf{Multiple Appropriate Facial Reaction Generation in Dyadic Interaction Settings: What, Why and How?}}


\author[1,2]{\fnm{Siyang} \sur{Song}}\email{ss2796@cam.ac.uk}
\equalcont{These authors contributed equally to this work.}

\author[1]{\fnm{Micol} \sur{Spitale}}\email{ms2871@cam.ac.uk}
\equalcont{These authors contributed equally to this work.}


\author[3]{\fnm{Yiming} \sur{Luo}}\email{yl1722@imperial.ac.uk}
\equalcont{These authors contributed equally to this work.}

\author[4]{\fnm{Batuhan} \sur{Bal}}\email{bal.batuhan@metu.edu.tr}

\author[1]{\fnm{Hatice} \sur{Gunes}}\email{hatice.gunes@cl.cam.ac.uk}

\affil[1]{\orgdiv{Department of Computer Science and Technology}, \orgname{University of Cambridge}}


\affil[2]{\orgdiv{School of Computing and Mathematical Sciences}, \orgname{University of Leicester}}


\affil[3]{\orgdiv{Department of Computing}, \orgname{Imperial College London}}

\affil[4]{\orgdiv{Department of Computer Engineering}, \orgname{Middle East Technical University}}





\abstract{According to the Stimulus Organism Response (SOR) theory, all human behavioral reactions are stimulated by context, where people will process the received stimulus and produce an \textit{appropriate} reaction. This implies that in a specific context for a given input stimulus, a person can react differently according to their internal state and other contextual factors. Analogously, in dyadic interactions, humans communicate using verbal and nonverbal cues, where a broad spectrum of listeners' non-verbal reactions might be \textit{appropriate} for responding to a specific speaker behaviour. There already exists a body of work that investigated the problem of automatically generating an appropriate reaction for a given input. However, none attempted to automatically generate multiple appropriate reactions in the context of dyadic interactions and evaluate the appropriateness of those reactions using objective measures. This paper starts by defining the facial Multiple Appropriate Reaction Generation (fMARG) task for the first time in the literature and proposes a new set of objective evaluation metrics to evaluate the appropriateness of the generated reactions. The paper subsequently introduces a framework to predict, generate, and evaluate multiple appropriate facial reactions.}




\maketitle

\section{Introduction}
\label{sec:Introduction}

According to the Stimulus Organism Response (SOR) model - proposed by Mehrabian and Russel \cite{mehrabian1974approach} - all behavioural responses or psychological changes in people are stimulated by their environment (or context), and people will inductively process the stimulus and modify their interactions to produce an appropriate response \cite{zhai2020sor, pandita2021psychological}, specific for each individual. The SOR model explains the relationship between stimuli (e.g., a speaker behavior) that have an impact on organisms internal states (e.g., listener cognitive and affective states) and the reaction that people generate to the stimuli (e.g., listener non-verbal behavior). Analogously, in dyadic interactions when humans communicate with each other using verbal and non-verbal cues, for a given input stimulus (i.e., from the speaker), a broad spectrum of responses (verbal) and reactions (non-verbal) might be \textit{appropriate} for an individual (i.e., listener) to generate according to their internal state. 

\begin{figure}
    \centering
    \includegraphics[width=\columnwidth]{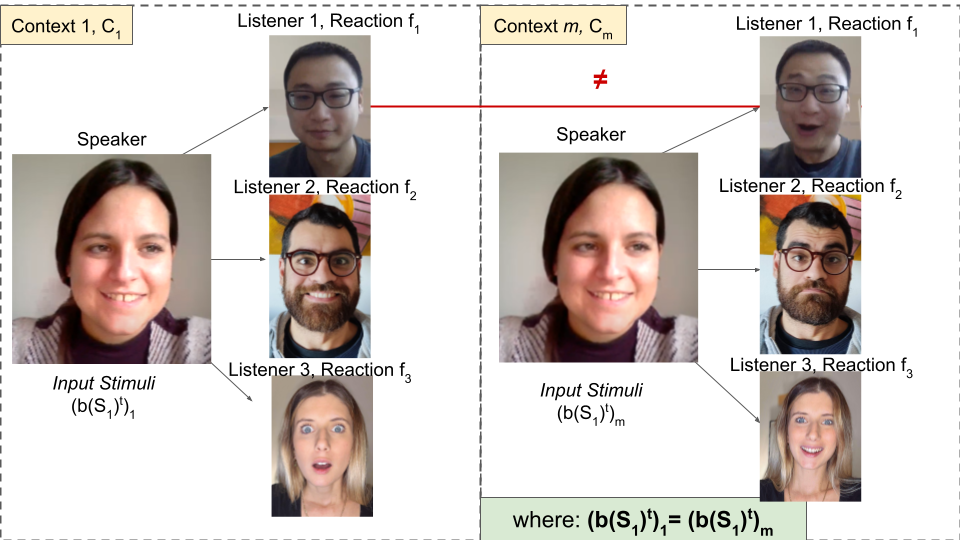}
    \caption{Multiple appropriate reaction generation in dyadic interaction settings. The same input stimuli ($(b(S_{1})^{t})_{1} = (b(S_{1})^{t})_{m}$) under different contexts ($C_{1} \neq  C_{m}$) of the speaker $S_{1}$ can elicit different reactions for the same listener ($(f_{1})_{1} \neq (f_{1})_{m}$) and also for different listeners ($f_{1} \neq  f_{2} \neq f_{3}$).}
    \label{fig:example}
\end{figure}

For example (see Figure \ref{fig:example}), during a conversation between an employee (Listener 1) and an employer (Speaker 1) who displays a specific behavior (or input stimuli, $b(S_{1})^{t}$) to compliment their employee  in a given context ($C_{1}$), the employee (Listener 1) might respond and react with much excitement ($f_{1}$). If the same information was communicated with the same behavior ($b(S_{1})^{t}$) but in a different context ($C_{m}$, where $(b(S_{1})^{t})_{1} = (b(S_{m})^{t})_{m}$), Listener 1's response and reaction ($f_{1}$) could differ significantly (e.g., with less excitement) depending on their contextual circumstances (e.g., having a particularly bad day), resulting in $(f_{1})_{1} \neq (f_{1})_{m}$. Analogously, another employee (Listener 2) would respond and react differently ($f_{2}$, e.g., with more excitement) than Listener 1 to the same stimuli produced by Speaker 1, resulting in $f_{1} \neq f_{2}$. As this example illustrates and the SOR theory suggests  \cite{mehrabian1974approach}, multiple responses and reactions are appropriate for a given input.

Recent years have witnessed an increasing number of studies targeting human-human dyadic interaction analysis, thanks to the wide application scenarios - such as, among others, surveillance, transportation, health, information security, and intelligent human agent interaction - and the advancement of pattern recognition, cognitive science, and neural networks \cite{peng2019affective}. 
Past works \cite{huang2017dyadgan, song2019exploiting, yoon2022genea,song2022learning,shao2021personality} have investigated the problem of automatically generating an appropriate response or reaction for a given input. Most of those studies focused on the generation of appropriate responses (e.g., using chat-bots \cite{song2019exploiting}) without considering the non-verbal reactions that enrich the message conveyed. Very few explored the generation of appropriate reactions via non-verbal behaviors \cite{yoon2022genea}, limiting their evaluation to a \textit{single appropriate} generated reaction - specifically hand gestures - via \textit{subjective measures}. 

As previously discussed, multiple reactions can be appropriate for the same context \cite{mehrabian1974approach}. However, none of the existing works attempted to automatically generate \textit{multiple appropriate reactions} in dyadic interaction settings and evaluate the appropriateness of those reactions using objective measures.
This paper defines the \textbf{Multiple Appropriate Reaction Generation} (MARG) task and proposes a new set of objective evaluation metrics to evaluate the appropriateness of the generated reactions for the first time in the literature. As a first step towards addressing the MARG research problem, it introduces a framework to predict, generate, and evaluate multiple appropriate \textit{facial} reactions in the same context. The facial MARG (fMARG) task specifically is a very challenging problem that has not been investigated yet due to multiple open research questions.

First, compared to standard machine learning and facial behavior analysis tasks \cite{li2020deep}, where each input data only correspond to one solution (e.g., a specific combination of action units corresponding to the emotion label of joy/happiness, 1-to-1 problem), the fMARG task is more difficult as it is a 1-to-N problem where each input stimulus may correspond to multiple appropriate facial reactions (research challenge 1, \textbf{RC1}). 
Second, most of the past works \cite{taras2020} explored the generation of an appropriate facial reaction proposing frameworks which only generate a single appropriate reaction. To the best of our knowledge, a framework to simulate and analyze the fMARG problem has not been defined yet, given the complexity and uncertainty of generating multiple appropriate facial reactions (research challenge 2, \textbf{RC2}). 
Third, the current available human-human dyadic interaction datasets \cite{palmero2021context,cafaro2017noxi,ringeval2013introducing} are designed to address various facial analysis tasks (e.g., facial affect recognition, personality recognition, etc.), but they are not labeled to provide a ground truth on appropriate facial reaction. Lastly, previous studies \cite{taras2020} measured and evaluated the appropriateness of generated reactions using subjective measures (i.e., running a user study, where participants evaluates the appropriateness of a specific reaction generated), which limit the reproducibility of these results. Currently, there is a lack of objective measures to evaluate the MARG tasks (research challenge 3, \textbf{RC3}).


This paper presents the first framework to predict, evaluate, and generate multiple appropriate facial reactions to address the above-mentioned open challenges. We present a model for generating multiple appropriate facial reactions in a specific context given an input (solution \textbf{S1}, addressing RC1). We create and design a theoretical framework to ground the modelling of facial MARG (\textbf{S2}, addressing RC2). Finally, we define a set of objective measures to evaluate the fMARG problem and to gain insights on how the proposed solutions can be improved (\textbf{S3}, addressing RC3).


\section{Hypotheses and task definition}
\label{subsec:task-definition}

\noindent This section formulates the hypothesis and formally defines the facial Multiple Appropriate Reactions Generation (fMARG) task.


\subsection{Hypotheses}

\noindent  \textbf{Hypotheses 1:} According to the SOR theory \cite{zhai2020sor, pandita2021psychological} and given the fuzzy nature of the fMARG problem, the same/similar behaviour expressed by a speaker could trigger different facial reactions expressed by not only various subjects but also the same subject under different contexts. Specifically, given a spatio-temporal behaviour $b_{S_n}^{t_1,t_2}$ expressed by a speaker $S_n$ at the time $[t_1,t_2]$, a set of (multiple) appropriate facial reactions $F_L(b_{S_n}^{t_1,t_2})$ could be expressed by different listeners, which can be represented as:
\begin{equation}
\begin{split}
&F_L(b(S_n)^{t_1,t_2}) =  \\ 
&\left [
\begin{array}{ccc}
     f_{L_1}(b_{S_n}^{t_1,t_2})_1 & \cdots & f_{L_1}(b_{S_n}^{t_1,t_2})_{I_1} \\
     f_{L_2}( b_{S_n}^{t_1,t_2})_1 & \cdots & f_{L_2}(b_{S_n}^{t_1,t_2})_{I_2}  \\
     \cdots &  \cdots & \cdots  \\
    f_{L_N}(b_{S_n}^{t_1,t_2})_1 & \cdots &  f_{L_N}(b_{S_n}^{t_1,t_2})_{I_N}
\end{array}
\right ]
\end{split}
\label{eq:hypothesis1}
\end{equation}
where $f_{L_\eta}(b_{S_n}^{t_1,t_2})_i$ ($i = 1, 2, \cdots, I_\eta$) denotes the appropriate facial reaction that could be expressed by the $\eta_{th}$ listener in response to the $b(S_n)^{t}$ under the $i_{th}$ context. Here, the number of possible appropriate facial reactions expressed by different listeners may not be the same (i.e., $I_1 \neq I_2 \cdots \neq I_N$). More importantly, the spatio-temporal patterns of these appropriate facial reactions are not guaranteed to be similar (illustrated in Figure \ref{fig:example}).


\noindent  \textbf{Hypotheses 2:} Listeners may express similar facial reactions in response to different speaker behaviours \cite{carminati2013effects}. For example, listeners may display similar positive facial expressions when a speaker tells either a joke or some good news. This means the $n_{th}$ listener's real facial reaction $f_{L_n}(b_{S_n}^{t_1,t_2})$ triggered by $b_{S_n}^{t_1,t_2}$ can also be an appropriate facial reaction in response to other speaker behaviours (e.g., the behaviour $b_{S_m}^{t_3,t_4}$ expressed by the $m_{th}$ speaker at the period $[t_3,t_4]$).


\subsection{fMARG task definition:}

\noindent Given the spatio-temporal behaviours $b_{S_n}^{t_1,t_2}$ expressed by the speaker $S_n$ at the period $[t_1,t_2]$, we define two types of fMARG tasks as follows:
\begin{itemize}
    \item \textbf{(i)} \textbf{Offline fMARG task:} this task aims to learn a ML model $\mathcal{H}$ that takes the entire speaker behaviour sequence $b_{S_n}^{t_1,t_2}$ as the input, and generates multiple appropriate spatio-temporal listener facial reaction sequences $P_f(b_{S_n}^{t_1,t_2}) = \{ p_f(b_{S_n}^{t_1,t_2})_1, \cdots, p_f(b_{S_n}^{t_1,t_2})_M \}$ as:
    \begin{equation}
    \begin{split}
    & P_f(b_{S_n}^{t_1,t_2}) = \mathcal{H}(b_{S_n}^{t_1,t_2})\\
    & p_f(b_{S_n}^{t_1,t_2})_1 \neq  p_f(b_{S_n}^{t_1,t_2})_2  \neq \cdots \neq p_f(L \vert b(S_n)^{t})_I \\
    & p_f(b_{S_n}^{t_1,t_2})_i \sim  f_{L_1}(b_{S_n}^{t_1,t_2})_i \in F_L(b(S_n)^{t_1,t_2})
    \end{split}
    \label{eq:task_define_1}
    \end{equation}
    where each $p_f(b_{S_n}^{t_1,t_2})_i \in P_f(b_{S_n}^{t_1,t_2})$ represents a generated facial reaction in response to $b_{S_n}^{t_1,t_2}$, which should be similar to at least one of the appropriate real facial reactions in $F_L(b(S_n)^{t_1,t_2}$ (defined in Eqa. \ref{eq:hypothesis1}) expressed by human listeners.

    \item \textbf{(ii)} \textbf{Online fMARG task:} this task aims to learn a ML model $\mathcal{H}$ that estimates each appropriate facial reaction frame (i.e.,  $\gamma_\text{th} \in [t_1,t_2]$ frame) by only considering the $\gamma_\text{th}$ frame and its previous frames expressed by the corresponding speaker (i.e., $b_{S_n}^{t_1,\gamma}$), rather than taking all $t_1 \text{th}$ to $t_2 \text{th}$ frames into consideration. This can be formulated as:
    \begin{equation}
        p_f(b_{S_n}^{t_1,t_2})_i^{\gamma} = \mathcal{H}(b_{S_n}^{t_1,\gamma})
    \end{equation}
    where $p_f(b_{S_n}^{t_1,t_2})_i^{\gamma}$ denotes the $\gamma_\text{th}$ predicted facial reaction frame of the $i_\text{th}$ generated appropriate facial reaction in response to $b_{S_n}^{t_1,t_2}$; and $b_{S_n}^{t_1,\gamma}$ denotes the speaker behaviour segment at the period $[t_1,\gamma]$. In summary, this task aims to gradually generate all facial reaction frames to form multiple appropriate spatio-temporal facial reactions as defined in Eqa \ref{eq:task_define_1}.

\end{itemize}

\section{Evaluation protocol} 
\label{subsec:evprot}

\begin{figure*}
    \centering
    \includegraphics[width =16cm]{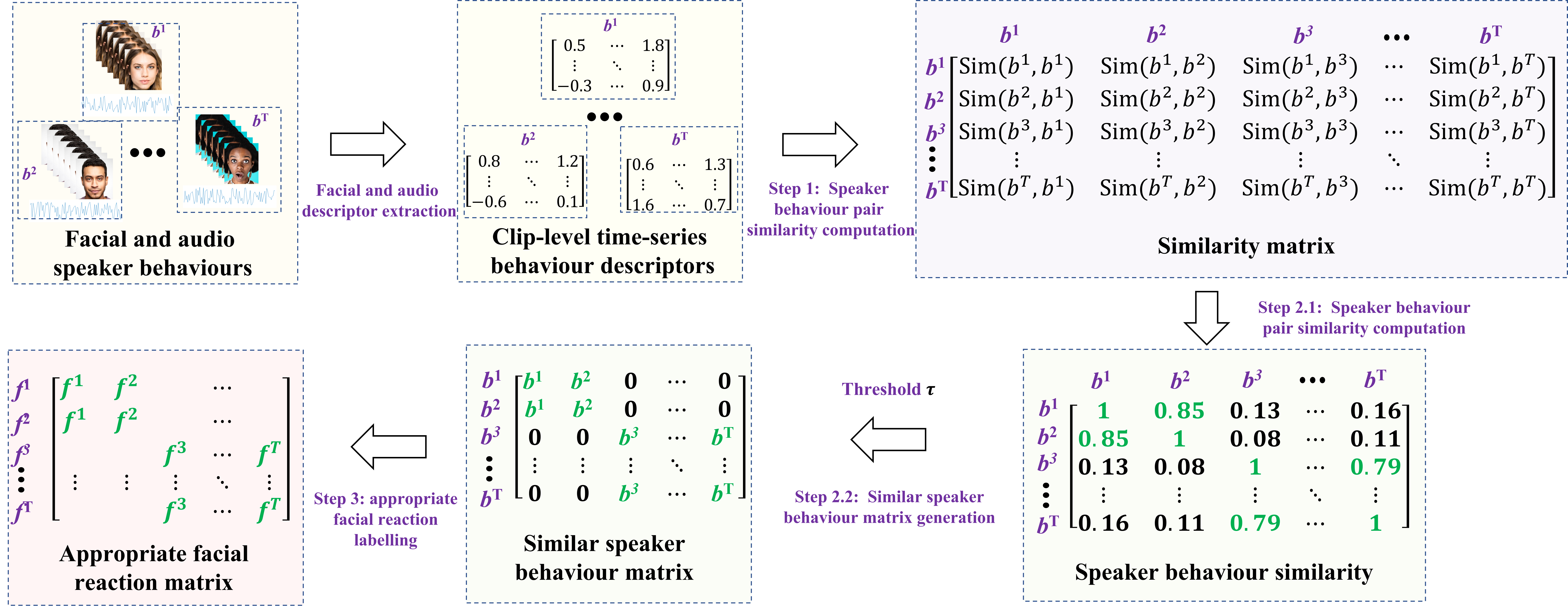}
    \caption{Illustration of the proposed automatic appropriate facial reaction labelling strategy (\textbf{Sec. \ref{subsec:appropriate_annotation}}). It first represents each audio-facial speaker behaviour clip as a multi-channel time-series signal. Then, we compute the similarity between each pair of speaker behaviour time-series (\textbf{Step 1}). After that, we define a set of similar speaker behaviours for each speaker behaviour (\textbf{Step 2}). Finally, for each speaker behaviour, we define the real facial reactions correspond to all of its similar speaker behaviours as its appropriate facial reactions (\textbf{Step 3}).}
    \label{Appropriate_definition}
\end{figure*}

\noindent As discussed in Sec. \ref{subsec:task-definition}, similar speaker behaviours may trigger listeners to express different facial reactions. Thus, given a speaker behaviour $b_{S_n}^{t_1,t_2}$, instead of only assessing the similarity between the generated facial reaction $p_f(b_{S_n}^{t_1,t_2})$ and the corresponding ground-truth (GT) real facial reaction $f_{L_n}^{t_1,t_2}$, we propose a set of objective evaluation metrics to evaluate the \textit{\textbf{appropriateness}}, \textit{\textbf{diversity}}, \textit{\textbf{realism}}, and \textit{\textbf{synchrony}} of the generated facial reactions $P_f(b_{S_n}^{t_1,t_2})$. To the best of our knowledge, this paper proposes the first set of objective evaluation metrics that assesses the appropriateness of the generated facial reactions.

\subsection{Automatic appropriate facial reaction labelling strategy}
\label{subsec:appropriate_annotation}

\noindent Since there is no golden standard for evaluating whether a facial reaction is appropriate in response to a speaker behaviour, while manually labelling would be labour-intensive and subjective, we first propose an objective automatic fMARG labelling strategy.

Given $T$ spatial-temporal behaviours $B(S) = \{b^1, b^2, \cdots, b^M\}$ expressed by a set of speakers, and $T$ GT real facial reactions $F(L) = \{f^1, f^2, \cdots, f^M\}$ expressed by the corresponding listeners, we propose to obtain all appropriate real facial reactions $F_L(b^m) = \{f^{m_1}_L, f^{m_2}_L, \cdots, f^{m_{I_m}}_L \} \in F(L)$ in response to each speaker behaviour $b^m \in B(S)$ as follows:
\begin{itemize}

     \item \textbf{Step 1:} We compute the similarities between $b^m$ and all speaker behaviours (including $b^m$ itself) in $B(S)$, resulting in $T$ similarity scores $\text{Sim}(b^m,b^1), \text{Sim}(b^m,b^2), \cdots, \text{Sim}(b^m,b^M)$.

    \item \textbf{Step 2:} We choose the speaker behaviours $B(S \vert b^m) = \{b^{m_1}, b^{m_2},\cdots, b^{m_{I_m}} \} \in B(S)$ as the \textit{similar speaker behaviours} of the $b^t$, which are defined by:
    \begin{equation}
        b^{\eta} \in B(S \vert b^m) \quad \text{Subjects to} \quad \text{SIM}(b^\eta,b^{t}) > \mathcal{T}
    \end{equation}
    where $\eta = 1, 2, \cdots, M$ and $\mathcal{T}$ is a threshold to decide whether the speaker behaviour $b^{\eta}$ is similar to $b^{t}$.
    

    \item \textbf{Step 3:} We define all real facial reactions corresponding to the speaker behaviours in $B(S \vert b^m)$ as the appropriate real facial reactions (defined as $F(L \vert b^m) = \{f^{m_1}, f^{m_2},\cdots, f^{m_{I_m}} \} \in F(L)$) in response to $b_m$, which can be formulated as:
    \begin{equation}
        f^\eta \in F(L \vert b^m) \quad \text{Subjects to} \quad b^\eta \in B(S \vert b^m)
    \end{equation}
    
\end{itemize}
In this paper, we extract three widely-used facial attribute descriptors, including $15$ facial action units (i.e., AU1, AU2, AU4, AU6, AU7, AU9, AU10, AU12, AU14, AU15, AU17, AU23, AU24, AU25 and AU26), facial affect - valence and arousal intensities - and the probabilities of eight categorical facial expressions (i.e., Neutral, Happy, Sad, Surprise, Fear, Disgust, Anger and Contempt), to represent frame-level human facial display. Specifically, all AUs' occurrences are predicted by the state-of-the-art GraphAU model \cite{luo2022learning,song2022gratis}, while facial affects and facial expression probabilities are predicted by \cite{toisoul2021estimation}). We also apply OpenSmile \cite{eyben2010opensmile} to extract clip-level audio descriptors, including GEMAP and MFCC features. Consequently, each speaker behaviour is represented by a multi-channel audio-visual time-series behavioural signal obtained by concatenating all frame-level descriptors.

Subsequently, we apply elastic similarity measurement \cite{shifaz2021elastic} (i.e., Dynamic Time Wrapping for multi-channel time-series) to compute the similarity between each pair of multi-channel time-series speaker behaviours, i.e., the function SIM is defined as:
\begin{equation}
    \text{SIM}(b^m,b^\eta) = 1 - \frac{\text{DTW}(b^m,b^\eta)}{\text{Max}_\text{DTW}}
\end{equation}
where $\text{DTW}(b_t,b_\eta)$ denotes the DTW distance between two multi-channel time-series representations of speaker behaviours $b_m$ and $b_\eta$; and $\text{Max}_\text{DTW}$ denotes the maximum DTW distance of all speaker behaviour pairs in the whole dataset.


\subsection{Evaluation metrics}
\label{subsec:metrics}


\noindent Given an well-trained ML model, we define $P_f^1, P_f^2, \cdots, P_f^M$ as the $M$  sets of real facial reactions generated based on the $M$ input speaker behaviours $b^1, b^2, \cdots, b^M$, where $P_f^m = \{p_f^{m_1} \neq p_f^{m_2} \neq \cdots \neq p_f^{m_\alpha} \}$ (i.e., we assume that the well-developed model can generate $\alpha$ different facial reactions in response to each speaker behaviour), and each input speaker behaviour $b_m$ corresponds to a set of appropriate real facial reactions $F(L \vert b^m)$. Then, we propose the following evaluation metrics to measure the performance of the developed model (i.e., the appropriateness, diversity, realism and synchrony of the generated facial reactions). In this paper, all speaker behaviours, real facial reactions and generated facial reactions are represented by multi-channel time-series facial attribute signals explained in Sec. \ref{subsec:appropriate_annotation}.


\subsubsection{Appropriateness metrics}

\noindent We first propose three metrics for evaluating the \textbf{appropriateness} of the generated facial reactions, including: (1) the DTW distance between the generated facial reaction and its most similar appropriate real facial reaction; (2) the Concordance Correlation Coefficient (CCC) between the generated facial reaction and its most similar appropriate facial reaction; and (3) Appropriate facial reaction prediction accuracy. Since DTW has been widely used to measure the similarity between two temporal sequences that may vary in speed \cite{rakthanmanon2012searching,li2020adaptively}, while the speeds for expressing similar facial behaviours may be varied across different subjects due to person-specific factors (e.g., age \cite{birren1995aging}), we propose to apply this metrics to measure the similarity between each generated facial reaction sequence and the real facial reaction sequence in our tasks. Meanwhile, the CCC has been frequently used to evaluate the correlation between human behaviour prediction sequence and the ground-truth sequence (e.g., dimensional affect recognition task \cite{zafeiriou2017aff,ringeval2013introducing}), and thus we also employ it as a key metrics:
\begin{itemize}

\item \textbf{(1) Facial reaction distance (\textbf{FRDist}):} we compute the DTW distance between the generated facial reaction and its most similar appropriate real facial reaction as:
\begin{equation}
 d_m = \sum_{i = 1}^{\alpha} \text{Min}(\text{DTW}(p^m_i, f^\eta_i))
\end{equation}
where $f^\eta_i \in F(L \vert b^m)$ denotes the most similar appropriate facial reaction to the generated $p^m_i$, and $d_m$ denotes the sum of the distances corresponding to all facial reactions generated in response to $b_m$. The final distance score for evaluation is obtained by averaging all obtained DTW distances:
\begin{equation}
 \text{FRDist} = \frac{\sum_{m = 1}^{M} d^m}{M}
\end{equation}

\item \textbf{(2) Facial reaction correlation (\textbf{FRCorr}):} we compute the correlation between each generated facial reaction and its most similar appropriate real facial reaction as:
\begin{equation}
 c_m = \sum_{i = 1}^{\alpha} \text{Max}(\text{CCC}(p^m_i, f^\eta_i))
\end{equation}
where $\text{CCC}$ denotes the Concordance Correlation Coefficient (CCC). The final correlation score for evaluation is obtained by averaging all obtained CCC values as:
\begin{equation}
 \text{FRCorr} = \frac{\sum_{m = 1}^{T} c_m}{M}
\end{equation}

\item \textbf{(3) Appropriate facial reaction prediction accuracy (\textbf{ACC}):}
\begin{equation}
    \text{ACC} = \frac{\sum_{m=1}^M \sum_{i=1}^\alpha \{p^m_i \in F(L \vert b^m) \}}{M \times \alpha}
\end{equation}
Here, the $p^m_i \in F(L \vert b^m)$ is conditioned on:
\begin{equation}
    \text{Sim}(p^m_i, f^\eta_i) > \mathcal{T}
\end{equation}
where the $\mathcal{T}$ is also employed as the threshold that decides whether the generated facial reaction $p^m_i$ is an appropriate facial reaction in response to $b_m$. It should be noticed that $p^m_i$'s most similar facial reaction in the entire dataset may not belong to the $F(L \vert b^m)$. In other words, we only evaluate the appropriateness by considering whether $p^m_i$ is similar to one of the appropriate real facial reactions defined by $F(L \vert b^m)$.

\end{itemize}

\subsubsection{Diversity, Realism and Synchrony metrics}

\noindent We expect the well-developed model can generate multiple different and photo-realistic appropriate facial reactions from each input speaker behaviour, as human listeners can express different reactions in response to the same speaker behaviour under different situations. 

Subsequently, we follow \cite{ng2022learning} to compute variation among all frames for evaluating the variance of each generated facial reaction. We also propose a metric called sum of Mean Square Error (S-MSE), to evaluate the diverseness among multiple generated facial reactions in response to the same speaker behaviour. In addition, a inter-condition diversity metric is also introduced to evaluate the diversity of all generated facial reactions in response to different speaker behaviours. These \textbf{Diversity} metrics are explained as follows:
\begin{itemize}

    \item \textbf{(1) Facial reaction variance (\textbf{FRVar}):} this aims to evaluate the variance of each generated facial reaction, which is obtained by computing the variation across all of its frames. The final facial reaction diversity is obtained by averaging the variance values of all generated facial reactions:
    \begin{equation}
        \text{FRVar}  = \frac{ \sum_{t = 1}^{M} \sum_{i = 1}^{\alpha} \text{var}(p^m_i)}{M \times \alpha}
    \end{equation}

    \item \textbf{(2) Diverseness among generated facial reactions (\textbf{FRDiv}):} we evaluate the model's capability in generating multiple different facial reactions by calculating the sum of the MSE among every pair of the generated facial reactions in response to each input speaker behaviour as:
     \begin{equation}
        \text{S-MSE}(m)  = \sum_{i = 1}^{\alpha - 1} \sum_{j = i + 1}^\alpha (p^m_i - p^m_j)^2
     \end{equation}
    The final diverseness score of the diverseness among generated facial reactions is obtained by averaging S-MSE scores produced from all input speaker behaviours as: 
     \begin{equation}
        \text{FRDiv}  = \frac{\sum_{m = 1}^{M} \text{S-MSE}(m)}{M}
     \end{equation}   
    
    \item \textbf{(3) Diversity among facial reactions generated from different speaker behaviours (\textbf{FRDvs}):} we finally evaluate the diversity of the generated facial reactions in response to different speaker behaviours as:
    \begin{equation}
        \text{FRDvs} = \frac{\sum_{i = 1}^{\alpha} \sum_{m = 1}^{M-1} \sum_{a = m+1}^{M} \text{MSE}(p^m_i, p^a_i)}{\alpha M(M-1)}
    \end{equation}

\end{itemize}

Then, we employ the Fréchet Inception Distance (FID) \cite{heusel2017gans} to evaluate the \textbf{Realism} the generated facial reactions, as it has been frequently employed for measuring the realism of the generated human facial and body behaviours in previous studies \cite{ng2022learning,li2021learn}:
\begin{itemize}

    \item \textbf{Facial reaction realism (\textbf{FRRea}):} the realism score is obtained by computing Fréchet Inception Distance (FID) between the distribution $\text{Dis}(P^m_f)$ of the generated facial reactions and the distribution $\text{Dis}(F(L \vert b^m))$ of the corresponding appropriate real facial reactions as:
    \begin{equation}
       \text{FRRea} = \text{FID}(\text{Dis}(P^m_f), \text{Dis}(F(L \vert b^t)))
    \end{equation}

\end{itemize}

Finally, we compute the Time Lagged Cross Correlation (TLCC) to evaluate the synchrony between the input speaker behaviour and the corresponding generated facial reaction behaviour, as this metrics has been frequently used to evaluate the leader-follower relationship \cite{boker2002windowed,ng2022learning}.
\begin{itemize}
    \item \textbf{Synchrony (FRSyn):} we first compute TLCC scores between the input speaker behaviour and each of its generated facial reaction as:
     \begin{equation}
        \text{Synchrony}(m)  = \sum_{i = 1}^{\alpha} \text{TLCC}(p^m_i, f^m_S)
     \end{equation}        
    where $f^m_S$ here denotes the multi-channel facial attributes time-series of the speaker behaviour $b_m$. Then, the final synchrony score is obtained by averaging synchrony scores obtained from all input speaker facial behaviour-generated facial reaction pairs:
     \begin{equation}
        \text{FRSyn}  = \frac{\sum_{m = 1}^{M} \text{Synchrony}(m)}{M}
     \end{equation}       
\end{itemize}

\section{Conclusion}

In this paper, we define a new affective computing research direction: Multiple Appropriate Facial Reaction Generation. We specifically present its basic theory and hypothesis, task definition, automatic appropriateness labelling strategy as well as a set of objective evaluation metrics.

\bibliographystyle{plain}
\bibliography{sn-bibliography}


\end{document}